\documentclass[times,twocolumn,final]{elsarticle}

\usepackage[export]{adjustbox}
\usepackage{amssymb}
\usepackage{latexsym}

\usepackage{url}
\usepackage{xcolor}

\usepackage{times}
\usepackage{graphicx}
\usepackage{amsmath}
\usepackage{amssymb}
\usepackage{algorithm}
\usepackage{tabularx}
\usepackage{verbatim}
\usepackage{array}
\usepackage{multirow}
\usepackage{amssymb}
\usepackage{mathtext}
\usepackage{colortbl}
\usepackage{musixtex}
\usepackage{mwe}
\usepackage{graphbox} 
\usepackage{multirow}
\usepackage{fancyhdr}
\usepackage{cuted}
\usepackage{latexsym}
\usepackage{mathtools}
\usepackage{breqn}
\usepackage{blindtext}
\usepackage{flushend}
\usepackage{algpseudocode}
\usepackage[caption=false]{subfig}

\makeatletter
\newcommand{\ALOOP}[1]{\ALC@it\algorithmicloop\ #1%
  \begin{ALC@loop}}
\newcommand{\ENDALOOP}{\end{ALC@loop}\ALC@it\algorithmicendloop}

\usepackage{algorithmicx}
\algdef{SE}[DOWHILE]{Do}{doWhile}{\algorithmicdo}[1]{\algorithmicwhile\ #1}%
\makeatother
\usepackage{mathtools}
 \newcommand{\sgn}{\operatorname{sgn}}

\journal{Journal of Image and Vision Computing}

\begin{document}

\begin{frontmatter}

\title{CrossATNet - A Novel Cross-Attention Based Framework for Sketch-Based Image Retrieval}


\author[mymainaddress]{Ushasi Chaudhuri \corref{mycorrespondingauthor}}
\ead{ushasi@iitb.ac.in}
\cortext[mycorrespondingauthor]{Corresponding author}

\author[mymainaddress,cminds]{Biplab Banerjee}

\author[mymainaddress]{Avik Bhattacharya}

\author[mysecondaryaddress]{Mihai Datcu}

\address[mymainaddress]{Centre of Studies in Resources Engineering (CSRE), Indian Institute of Technology Bombay, India.}
\address[cminds]{Center for Machine Intelligence and Data Science (C-MInDS), Indian Institute of Technology Bombay, India.}
\address[mysecondaryaddress]{Deutsches Zentrum für Luft- und Raumfahrt (DLR), Oberpfaffenhofen, Germany.}

\begin{abstract}
We propose a novel framework for cross-modal zero-shot learning (ZSL) in the context of sketch-based image retrieval (SBIR). Conventionally, the SBIR schema mainly considers simultaneous mappings among the two image views and the semantic side information. Therefore, it is desirable to consider fine-grained classes mainly in the sketch domain using highly discriminative and semantically rich feature space. However, the existing deep generative modelling based SBIR approaches majorly focus on bridging the gaps between the seen and unseen classes by generating pseudo-unseen-class samples. Besides, violating the ZSL protocol by not utilizing any unseen-class information during training, such techniques do not pay explicit attention to modelling the discriminative nature of the shared space. Also, we note that learning a unified feature space for both the multi-view visual data is a tedious task considering the significant domain difference between sketches and the colour images. In this respect, as a remedy, we introduce a novel framework for zero-shot SBIR. While we define a cross-modal triplet loss to ensure the discriminative nature of the shared space, an innovative cross-modal attention learning strategy is also proposed to guide feature extraction from the image domain exploiting information from the respective sketch counterpart. In order to preserve the semantic consistency of the shared space, we consider a graph CNN based module which propagates the semantic class topology to the shared space. To ensure an improved response time during inference, we further explore the possibility of representing the shared space in terms of hash-codes. Experimental results obtained on the benchmark TU-Berlin and the Sketchy datasets confirm the superiority of CrossATNet in yielding the state-of-the-art results.
\end{abstract}


\end{frontmatter}


\section{Introduction}
The recent years have witnessed the accumulation of a vast volume of data, thanks to the availability of different types of comparatively cost-effective sensors. As a result, a given phenomenon can be realized in various forms of representations, thus providing complementary perspectives. Given the inherent distributions shift among varied data modalities, it is non-trivial to jointly analyze them within the scope of the original data space. The modelling of a shared embedding space is necessary for multi-modal data projection, preferably in a discriminative manner. In this paper, we are particularly interested in the task of cross-modal image retrieval. This task is essential since the visual data can be acquired in different ways; for example, colour image, depth image, contour or sketch image, etc. As far as the potential applications are concerned, retrieving depth from the concerned RGB image~\cite{torralba2002depth} for path-planning in SLAM models is a sought-after problem in this regard. 


Amongst others, the notion of \textit{Sketch-based image retrieval} (SBIR) is nowadays prevalent since it is easy to obtain a rough sketch drawing for any object type. The {\it mental target}~\cite{xu2020mental} or query sample can be drawn by any user in the absence of image instances of some obsolete classes. Beside, sketches are a highly symbolic and minimalistic representation of data. In particular, SBIR finds prominent applications in forensic studies or police investigations. By definition, SBIR systems seek to retrieve several related colour images for a given query sketch image. Often it is easier to quickly draw a sketch instance of a query image than to find its actual data instance. The problem set is hugely demanding, primarily since the sketch and colour images are inherently diverse in appearance and content. While their spectral and textural properties characterize natural images, shape plays the most crucial role in understanding the sketch images. In the same light, recent studies~\cite{geirhos2018imagenet} have shown that the ImageNet-trained convolutional neural networks (CNNs) have an inherent bias towards texture. Hence, such CNN cannot be directly applied to extract meaningful features from the sketch images. Instead, we need a sophisticated model that is deemed to increase the shape bias to improve the overall representation, given sketch images.

\begin{figure}
    \centering
    \includegraphics[width=\linewidth]{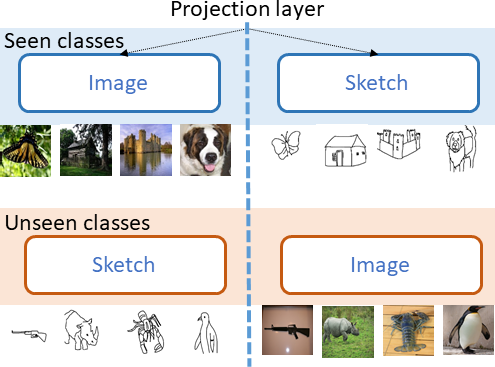}
    \caption{Overview of the proposed CrossATNet model which allows a cross-modal data retrieval, by reducing the distances of each mode of data from a shared embedding vector. }
    \label{fig:smallblock}
\end{figure}

Typically, the existing cross-modal retrieval systems (including SBIR) work in a supervised setting where the same set of semantic categories are utilized during training and inference. However, given the diversity of the real-world objects, the retrieval model trained on a given set of classes should generalize well to previously unseen categories with bare minimum side information. This condition increases the challenge of this problem a notch higher. On a different note, zero-shot learning (ZSL) has recently been extensively explored in image recognition. The ZSL models are generally trained on a given set of \textit{seen} classes. The system is required to properly categorize samples coming from an entirely non-overlapping collection of \textit{unseen} classes during the inference phase (figure~\ref{fig:smallblock}). In addition to the visual instances, semantic class-prototypes (one per class) are available for both the seen and unseen data. In this regard, the ZSL problem can be devised as a many to one mapping problem from the visual to the prototype space given the seen classes.  It can then be utilized as the mapping function for classifying the unseen class data. The ZSL paradigm can be incorporated within the SBIR framework as well. However, it has to be ensured that the mapping functions from both the visual modalities (colour and sketch images) to the semantic space should be modelled appropriately.

Very recently, the ZSL based SBIR systems are developed mainly in conjunction with the deep generative models~\cite{shen2018zero,kiran2018zero}. In this technique, given a sketch sample, the authors try to learn the corresponding class RGB image by generating pseudo samples corresponding to the unseen class prototypes. This requires access to unseen class samples during the training time. This fundamentally violates the protocol of zero-shot learning.  They pose the ZSL-SBIR problem as a standard supervised learning problem of deploying the notion of adversarial learning to align the cross-modal visual space to the semantic space. Also, adversarial training can be unstable if the min-max problem is not intuitively designed. The pseudo unseen sample generation technique may fail to generate unseen samples which exactly overlap with the original unseen data in some feature space. On the other hand, we are keener on making the shared feature space discriminative, learning relevant image features given the sketch data, and avoiding domain dependence of the shared space. We bridge the domain gap by imposing semantic consistency by preserving the topology of the semantic labels in the network. If such a space can be realized during training, we can expect the unseen test classes also to follow the same and yield excellent results upon deployment in the testing phase.  

To alleviate these shortcomings, we propose an encoder-decoder based deep representation learning strategy for zero-shot cross-modal retrieval problem. In the proposed framework, we use sketch-anchored and image-anchored cross-modal triplets to align sketches and images in the semantic space. We also recommend a novel cross-attention network wherein we use a sketch-guided feature extraction strategy from images. This helps in approximating the complicated image-sketch mapping-function by enabling focus on specific parts of the input image. To reduce the domain gap between the two modalities, we preserve the topography of the semantic labels in the embedding space. This reduces the confusion of assigning incorrect attribute information to a cluster during the inference phase. The semantic information is fed to the network by forming a semantic topology graph and injecting the graph convolution network (GCN) features to the semantic vector. Finally, to incorporate a speedy retrieval framework, we use a hashing function to convert the shared feature space to ternary bits. This increases the efficiency of the network. In the following, we summarize the significant highlights of the proposed models:


1) We introduce a novel CrossATNet framework which constructs a shared discriminative features space for efficient sketch-based image retrieval. 2) Some of the highlighting aspects of CrossATNet are, a) cross-triplet, b) cross-attention, and c) ensuring semantic consistency. 3) We consider both real-valued and hash-code based shared space to analyze the trade-off between accuracy and retrieval time. 4) Results on Sketchy and TU Berlin are promising, and the robustness of the model is tested with several ablation studies.

\vspace{-3mm}
\section{Related Works}
Considering the primary attention of our paper, we briefly discuss the relevant emerging works in image retrieval in-line with (i) cross-modal data and (ii) zero-shot learning. 

\textbf{Retrieval from cross-modal data:} Some of the early works in data retrieval was done by \cite{ferecatu2004retrieval,gavat2007knowledge} in single modal image retrieval. However, in recent times, the focus has shifted to more challenging cross-modal data retrieval tasks given its wide applications in different domains. Most of these researches are based on image-text cross-domain retrieval \cite{mandal2017generalized}. There are also a few works in the image to its depth retrieval \cite{torralba2002depth}, gaining its popularity primarily after the LiDAR point cloud data became available. Another attempt in this line has been for retrieving image-speech cross-modal data \cite{cmirnet}. Efforts for solving SBIR has also gained motion after the \textit{sketchy} database came up. \cite{zhang2018generative,zhang2016zero,yang2016revisiting,kodirov2017semantic} are a few of the notable works in this line. However, it can be noted that all these algorithms were designed for SBIR, and not the vice versa. Interestingly, many researchers have extended retrieval techniques to irregular domains using graph convolution networks \cite{neurocomputing,cviu}, and point-CNN based techniques \cite{li2018pointcnn}. 

\textbf{Zero-shot learning (ZSL):} Zero-shot learning is being able to solve a task, despite not having any training example of that class, and just by using some semantic information. \cite{fu2018recent} provides a complete comprehensive study in the recent advances in zero-shot learning problems. One of the early notable work was by \cite{akata2015evaluation}. The ZSL concept mainly gained attention from this work, and since then, it has been studied extensively.
In ZSL, usually, some lateral information is required to transfer the knowledge learned in the seen classes to the unseen classes. Often, this lateral information is realized by visual semantic mappings~\cite{gune2018structure}. People have also come up with understanding the attribute mappings in a semantic space, which can be manually defined~\cite{farhadi2009describing}, or by using Word2Vec embeddings~\cite{dutta2019semantically}, or by using a sentence description~\cite{reed2016learning}. Currently, the GAN based architecture is the commonly used framework for ZSL. Here, given an unseen sample and its prototype, the network generates a pseudo-unseen sample of a different mode of data \cite{kumar2018generalized}. Presently, various detection and cross-modal retrieval frameworks are achieved, apart from the classification task in ZSL.~\cite{parida2020coordinated} proposed a coordinated joint multi-modal embedding space for audio-visual classification and retrieval of videos, using ZSL. 

To the best of our knowledge~\cite{kiran2018zero,shen2018zero,dutta2019semantically,sake2019} are presently the state-of-the-art methodologies that exist in the literature for zero-shot SBIR. It may be noted that while~\cite{kiran2018zero,dutta2019semantically} use a generative adversarial network (GAN) based approach for this task,~\cite{shen2018zero} use a graph-convolution network for aligning sketches and images in the shared semantic space. The authors in~\cite{sake2019} advocate the importance of domain adaptation by transferring the knowledge acquired from ImageNet to their sketch data to improve the model's transferability. In another study,~\cite{pandey2020stacked} also came up with a stacked adversarial-based network for ZSL SBIR. However, as of now,~\cite{dutta2019semantically} and~\cite{sake2019} remain the state-of-the-art architecture in ZSL: SBIR, and hence have been used for comparison.  

We want to emphasize that our model is a robust cross-modal retrieval framework in contrast to the few existing methods on sketch-based image retrieval. Furthermore, It also encodes the image-based sketch retrieval information. Current works in SBIR~\cite{dutta2019semantically,sake2019} exploit a generative framework which typically requires sketch samples to learn the corresponding class RGB image by generating pseudo samples. This procedure requires access to unseen class samples during the training time. Therefore, it fundamentally violates the protocol of zero-shot learning. Our model is an encoder-decoder framework where we use the concept of cross-triplets for aligning the image-sketch modality in the feature space. We also preserve the topology of the semantic space using a minimally-connected graph, which helps in bridging the domain gap between the two modalities. We also introduce a novel cross-attention framework to extract a sketch-guided feature extraction from images to handle the classes which have highly-cluttered features.


\section{Methodology}
\noindent \textbf{Preliminaries:} Let $\mathcal{X}$ and $\mathcal{Y}$ denote the two incoming input streams to the network corresponding to sketches and images, respectively. Under this setup, we aim to retrieve the top-$k$ images from $\mathcal{X}$/$\mathcal{Y}$, given a query image from the modality $\mathcal{Y}$/$\mathcal{X}$. For this purpose, we subdivide the data into train and test classes, namely the seen and the unseen classes as defined previously in such a way that $\mathcal{X} = \{\mathcal{X}^{s} \cup \mathcal{X}^{u}\}$. Here $s$ denotes the seen training classes and $u$ denotes the unseen test classes, We impose the strict constraint that $\mathcal{X}^{s} \cap \mathcal{X}^{u} = \varnothing$. Similarly, we also subdivide the $\mathcal{Y}$ data into the seen and the unseen data classes. Let us define the label set for the data classes as $\mathcal{Z}$ and the semantic class prototypes for the same as $\mathcal{W}$. The semantic prototypes are a unique description of each class.

\begin{figure*}
    \centering
    \includegraphics[width=0.75\linewidth]{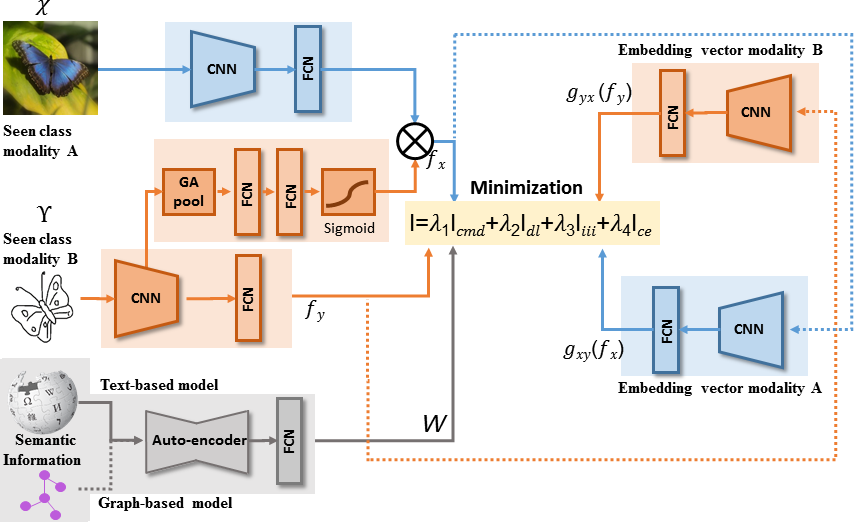}
    \caption{Overall pipeline of the proposed CrossATNet architecture for a zero-shot retrieval from a cross-modal database, by projecting the data samples on a shared semantic space. The network uses cross-attention to extract sketch-guided features from images. The network is trained on the \textit{seen} classses by forming cross-triplets, and tested on \textit{unseen} classes. With the help of the projection layer, retrieval from unseen class are possible.}
    \label{fig:block}
\end{figure*}

The model is trained using only the seen instances, while the unseen instances are deployed during the inference phase. In this work, we primarily aim to design a domain-agnostic mapping between the different data modalities in the visual space and the semantic space. 
The overall architecture of our proposed framework is shown in terms of a block diagram in figure~\ref{fig:block}.

\subsection{Overall CrossAT:Net Architecture}
We use a two-stage training process as described below to design our framework.
\begin{enumerate}
    \item We first extract the primary level features from the image and sketch instances by training modality-specific classifiers.
  
    \item Next, we carry out the training of the CrossAtNet framework, which primarily consists of an encoder-decoder-based model to carry out the visual space to semantic space mapping using a cross-attention network.  
\end{enumerate}

To extract the primary level features for both the data modalities, we perform transfer-learning from a standard pre-trained network, trained on the Imagenet dataset. We then fine-tune them as per our dataset and its classes for both the modalities $\mathcal{X}^s$ and $\mathcal{Y}^s$. Both the modality-specific classifiers are trained separately to preserve the domain-specific information. For this purpose, we use the VGGNet and the ResNet pre-trained networks for the initial weight-initialization for the subsequent stage on $(\mathcal{X}^s$ and $\mathcal{Y}^s)$, using the labels from $\mathcal{Z}^s$.

We initialize the second stage by the network weights obtained from stage I. We aim to learn the shared latent space features of both the modalities trained. The model should be trained so well on the seen classes that they perform equally well on unseen class instances when deployed during the inference phase. The latent shared features space denoted as \textbf{V} should be designed in a way such that it data instances from $\mathcal{X}^s$ and $\mathcal{Y}^s$ under similar classes should be closely aligned to their corresponding semantic information feature $\mathcal{W}^s$. The proposed framework has mainly four branches: visual encoders, semantic encoder, cross-attention network, and a decoder network. The architectures of these branches are described in the following subsections.

\subsubsection{Visual Encoders:}
For the visual data inputs from the streams $\mathcal{X}^s$ and $\mathcal{Y}^s$ we use encoder functions $f_x(\theta_x)$ and $f_y(\theta_y)$, where $\theta_x$ and $\theta_y$ are the learnable parameters for the sketch and the image data streams, respectively. 

To extract the image features, we use a sketch-guided cross-attention network. To realize this, the input weights received from the stage I from the sketch module $\mathcal{X}$ is branched, with one branch fed to the encoding function  $f_x(\theta_x)$, and the other branch is connected to the cross-attention branch (encoded by $f_{\text att}(\theta_{\text att})$). Again, $\theta_{\text att}$ here is the learnable parameters for this set of encoding. For realizing the cross-attention network, we take the inputs from stage I and use a {\tt global-average pooling} layer. This is then followed by 2-layers of fully-connected layers. The output of this is finally passed through a Sigmoid function. This entire architecture constitutes the cross-attention network module. The output of this encoder is then directly multiplied with the output from the image encoder ($f_y(\theta_y)$).

\subsubsection{Semantic Encoders:}

\begin{figure}
    \centering
    \includegraphics[width=0.6\linewidth]{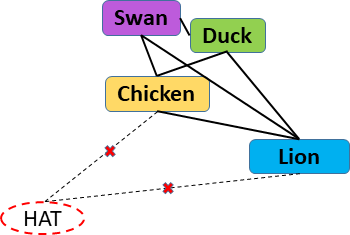}
    \caption{A hierarchy-graph structure by constructing minimum spanning tree of the Word2Vec features of the labels.} 
    \label{fig:graph}
\end{figure}

The semantic information is a unique description of each class label. So the total number of different semantic vector presents would be the number of training classes. We encode the semantic information similar to the visual encoder part by using the mapping function $f_w(\theta_w)$, where $\theta_w$ is its corresponding set of learnable parameters. In most of the existing work, a distributed word-vector embedding vector is used for each class name. Most commonly, this is achieved by using the standard { \tt word2vec} encoding. The word2vec module is pre-trained on large natural language databases, which leads to the preservation of the English language word topography in the semantic space. For example, the embedding vector for the word {\tt cat} should be closer to that of {\tt dog} in the semantic space, than it would be with that of {\tt Airplane}. Figure~\ref{fig:graph} illustrates this preservation of the semantic topography within different words through an example. We take the semantic encoding from two network branches, i) word2vec encoding, and ii) semantic-topography graph preservation. For the latter part, we create a minimally-connected graph of the nearest topographic classes and use a graph-convolution network to propagate this information onto the latent shared space. Preserving the topology of the classes in the network helps in reducing the domain-gap between the different modalities by aligning the classes of the two domains in a similar way. The encoder function $f_w(\theta_w)$ has a series of fully-connected layers, which basically encodes the projections from the word2vec module and the graph module together in the semantic space.

However, it is not ensured whether this semantic information is directly propagated in the shared latent space given the non-linear mappings performed at each neuron. 
This is a severe obstacle in ZSL where the unseen samples are mapped trivially in the latent space, considering that i) model is trained only on the seen classes, ii) there is significant distributions difference between the seen and unseen visual samples. Hence, we consider the following two different design protocols for the semantic branch.
\begin{itemize}
    \item The semantic encoder is defined as a series of fully-connected layers to project the seen class prototypes $\mathcal{W}^s$ onto the shared feature space. However, in this case, the original semantic space topology may not be preserved in the latent space.
    \item We use the word2vec features along with the graph $\mathcal{G}$, encoding the neighborhood topology of $\mathcal{Z}^s $, to form the semantic vector $\mathcal{W}^s$. We use a graph convolution network (with graph convolution and graph pooling layers) to encode $\mathcal{G}$, and a fully-connected layer to fuse the word2vec and graph features to form the semantic vector $\mathcal{W}^s$.
\end{itemize}

\subsubsection{Cross-modal reconstruction networks}
We use two reconstruction branches which reconstruct the cross-modal data instances from a given modal data. We define this decoding network with the decoder functions  $g_{xy}(\theta_{xy})$ and $g_{yx}(\theta_{yx})$. When we provide the feature embedding of the sketch data, i.e $f_x(X^s_c)$ and encode it using the decoder network $g_{xy}(f_x(X^s_c))$, we get the feature embedding of the corresponding label image data instance, $f_y(Y^s_c)$. Here, $c$ denotes a particular label class. Hence, given the feature encoding of one stream, we can reconstruct the encoding of the complimentary stream. The decoder network effectively acts as a regularizer function and helps in achieving domain invariance in the shared latent space.

\subsection{Hashing}
The shared space that has been realized is real-valued. This requires Euclidean distance measures for searching during the inference phase. If we transform the shared features into a ternary space, we can use hamming distance measures for speedy retrieval. The performance of a model is predominantly dependant on two things: accuracy and efficiency. While it is crucial to improve the accuracy of a model, we should not completely ignore the efficiency aspect. Hence in this regard, we add a hashing module that encodes the embedding space into ternary bit strings, which comprises of -1, 0, and 1. This helps in attaining a higher efficiency of the model as the comparisons can be performed by using simple XOR operations during retrieval. We use the standard Signum function, as defined in equation~\ref{eq:hash} for hashing, where X is the concerned modality of the input data stream. We denote the learned transformations in this step as $W_t$. We refer to this learned shared space, subjected to the hashing function as \textbf{H}. The hashing function hence transforms the shared feature space from \textbf{V} to \textbf{H}.
\begin{equation}\label{eq:hash}
    \textbf{H} = S_X = \sgn(W_t \times f_{X}(\mathcal{X}) )
\end{equation}

\subsection{Objective Function}
In order to realize the latent space $\textbf{H}$, we introduce the following loss functions to the network.

\noindent \textbf{a) Cross-modal latent loss ($\mathcal{L}_{cmd}$):} It is desirable to get the visual encoding of both the data streams adjacent in the shared feature space by projecting them onto a common entity. To achieve this, we reduce the distance between the visual encoding of both the data of the same class $(x_i^s,y_i^s)$ to the semantic vector encoding of the same class $w_i^s$. We reduce the distance by minimizing the mean-square error (MSE) between them. This procedure mainly helps in bringing down the cross-modal intra-class variance. We define this loss function as,
\begin{equation}\label{eq:cmd}
    \centering
    \mathcal{L}_{cmd} =  ||f_x(\mathcal{X}^s_c) - f_w(\mathcal{W}^s_c)||_{\mathbf{F}}^2 + ||f_y(\mathcal{Y}^s_c) - f_w(\mathcal{W}^s_c)||_{\mathbf{F}}^2 
\end{equation}

Here, \textbf{F} represents the Frobenious norm, defined as the square root of the sum of the absolute squares of the matrix elements.

\noindent \textbf{b) Cross-triplet loss ($\mathcal{L}_{iii}$): } We add a cross-triplet loss to reduce the intra-modal distances while increasing the inter-class distances in the shared feature space. To extend this loss to a cross-modal setup, we use two types of triads. One set of triads are constructed by taking an image as an anchor and using the same class sketch instance and a different class sketch instance. This scheme is described by the equation~\ref{eq:3a}. The other set of triads are formed by taking a sketch instance as an anchor and its same corresponding class and other class image instance along with it. This set of triads are described by equation~\ref{eq:3b}.
\begin{equation}\label{eq:3a}
   \mathcal{L}_{3a} =\operatorname{max} \left (
 { d(\operatorname{f}_y \left ( \mathcal{Y}^s_c \right ), \operatorname{f}_x \left ( \mathcal{X}^s_c \right ))}
 - {d( \operatorname{f}_y \left ( \mathcal{Y}^s_c \right ),\operatorname{f}_x \left ({\mathcal{\tilde{X}}^s_c} \right ))}
 + \alpha, 0 \right )
\end{equation}
\begin{equation}\label{eq:3b}
     \mathcal{L}_{3b} =\operatorname{max} \left (
 { d(\operatorname{f}_x \left ( \mathcal{X}^s_c \right ), \operatorname{f}_y \left ( \mathcal{Y}^s_c \right ))}
 - {d( \operatorname{f}_x \left ( \mathcal{X}^s_c \right ),\operatorname{f}_y \left ({\mathcal{\tilde{Y}}^s_c} \right ))}
 + \alpha, 0 \right )
\end{equation}

Here, we use $\mathcal{\tilde{Y}}^s_c$ to denote a seen image instance of any class other than $c$ and $d()$ to denote the Euclidean distance between any two vectors. $\alpha$ denotes the heuristically chosen margin value, which pushes apart the different classes beyond that minimum distance. The image-anchored $\mathcal{L}_{3a}$ loss and the sketch-anchored $\mathcal{L}_{3b}$ loss together comprises of the cross-triplet loss $\mathcal{L}_{iii}$.

\noindent \textbf{c) Decoder loss ($\mathcal{L}_{dl}$):} To make the shared features domain-agnostic, we use a decoder loss in the reconstruction sub-branch of the network. To achieve this, we take the visual encoding of a data instance from a particular modality and reconstruct the visual encoding of instances from the alternate modality (of the same class $c$). This helps us in achieving domain-invariance in the shared feature space as we taper their distribution-gap. The idea is to make the out of the decoder function  $g_{xy}(f_x(\mathcal{X}^s_c))$ near-equal to $f_y(\mathcal{Y}^s_c)$ (similarly for the other decoder function), hence reducing the distribution-gap between the different modalities. The sketch and the image data vary considerably in the visual feature space due to the absence of texture information in sketches. This loss helps in class-wise aligning the two different modalities better in the decoder space. The loss function is defined in equation~\ref{eq:dl}.
\begin{equation}\label{eq:dl}
    \mathcal{L}_{dl} =||g_{xy}(f_x(\mathcal{X}^s_c)) - f_y(\mathcal{Y}^s_c)||_{\mathbf{F}}^2 +  ||g_{YX}(f_y(\mathcal{Y}^s_c)) - f_x(\mathcal{X}^s_c)||_{\mathbf{F}}^2
\end{equation}

\noindent \textbf{ {Classification loss ($\mathcal{L}_{ce}$):}} Finally, to make the shared feature space sufficiently discriminative for both the modalities, we inject the class information. This helps in increasing the inter-class distances in the visual encoding space. We use the standard cross-entropy function as defined in equation~\ref{eq:ce}. We take the cross-entropy loss of the hashed encoding to make the hashed space sufficiently discriminative.
\begin{equation}\label{eq:ce}
    \centering
    \mathcal{L}_{ce} =\text{CE} \left (S_A \right ) +  \text{CE} \left (S_B  \right )
\end{equation}

The overall objective function is the weighted sum of all the loss functions, as mentioned above. The weights act as momentum in the network. The final objective function for our proposed framework is given in equation~\ref{eq:total}.

\begin{equation}\label{eq:total}
    \centering
    \mathcal{L} = \lambda_1 \mathcal{L}_{cmd} + \lambda_2 \mathcal{L}_{dl} + \lambda_3 \mathcal{L}_{iii} + \lambda_4 \mathcal{L}_{ce}
\end{equation}

The presence of multiple loss functions transforms the problem into a non-convex optimization problem. However, when each loss is considered individually for the optimization while keeping all the other losses constant, the problems transform back to a convex optimization problem for that loss. Likewise, if we iteratively perform an iterative-shrinkage gradient descent for each loss individually, the network can be trained efficiently.

\subsection{Inference} 
\noindent \textbf{Cross-modal retrieval:} Once the network is trained, we get the shared feature space \textbf{H}. To find the top-$k$ retrieved data instances of a query $q$, we first find its representation in the shared embedding space and search for its $k$-nearest neighbour ($k$-NN) feature instances using Euclidean or Hamming distance, depending on the variant of our architecture. These are the top-$k$ retrieved data instances. For a query $x_i^u$ from $\mathcal{X}$, we find the top-$k$ retrieved data instances from $\mathcal{Y}$, and vice versa. 

\noindent \textbf{Unimodal retrieval:} Likewise, for a query $x_i^u$ from $\mathcal{X}$, we find the top-$k$ retrieved data instances from $\mathcal{X}$, and vice versa. The overall framework of the proposed architecture is demonstrated in algorithm~\ref{algo:framework}.

\begin{algorithm}
\caption{The proposed training and inference stage}\label{algo:framework}
\begin{algorithmic}[1]

	\Require $ \{\mathcal{X}^s, \mathcal{Y}^s, \mathcal{W}^s, \mathcal{Z}^s\}$
	\Ensure Unified representations $\bf{H}$. 
	\State \textbf{Stage 1:} Normalize and pre-train $\mathcal{X}^s$ and $\mathcal{Y}^s$.
	\State \textbf{Stage 2:} Find the Word2Vec embeddings of $\mathcal{Z}^s$.
	\State Construct a topography graph $\mathcal{G}$ and generate $\mathcal{W}^s$. 
	\Do
		\State \begin{equation} \underset{\theta_x, \theta_y,\theta_{xy}, \theta_{yx}}{\min}\lambda_1 \mathcal{L}_{cmd} + \lambda_2 \mathcal{L}_{dl} + \lambda_3 \mathcal{L}_{iii} + \lambda_4 \mathcal{L}_{ce}  \end{equation}
	\doWhile {until loss convergence}\\
	\Return  \textbf{$S_X$} and \textbf{$S_Y$} (Projection from \textbf{H})
	\vspace{2mm}
	\hrule
	\vspace{2mm}
	\Require $x^u_i \in \mathcal{X}^u$ or $y^u_i \in \mathcal{Y}^u$ and $\mathcal{W}^u$
	\Ensure Top-$K$ retrieved data.
	\State Cross-modal zero-shot retrieval using $k$-NN.
\end{algorithmic}
\end{algorithm}


\section{Experiments}
\noindent \textbf{Datasets: } We validate our model by performing the experiments on two standard benchmark datasets: Sketchy~\cite{sangkloy2016sketchy} and TU-Berlin~\cite{liu2017deep} dataset. 

The Sketchy dataset~\cite{sangkloy2016sketchy} is a collection of a large number of sketches and images of the same class labels. Each image/sketch instance has multiple corresponding sketches/images. There are 125 different classes. Each class has 100 images, making it a total of 12,500 images. Similarly, for the sketches, there are multiple sketchers for each image. So the number of sketches vary in each class. There are a total of 75,471 sketches distributed over 125 categories. For our ZSL framework, we randomly select 25 classes as the unseen classes for the inference phase and use the rest for the training purpose as the training dataset. 

\begin{table*}\caption{Performance of the proposed CrossATNet framework for sketch-based image retrieval on the \textbf{Sketchy} and \textbf{TU-Berlin} dataset. Here, $\dagger$ represents the algorithms that have a hashed vector-representation. }\label{tab:sketchyberlin} 
\scalebox{0.80}{
\begin{minipage}{20cm}
\begin{center}
\begin{tabular}{@{\extracolsep{10pt}}l l llllc@{}}
    \hline
    \multirow{1}{*}{} &
    \multirow{1}{*}{\textbf{Task}} &
    \multicolumn{2}{c}{\textbf{Sketchy}} &
    \multicolumn{2}{c}{\textbf{TU-Berlin}}&
    \multicolumn{1}{c}{\textbf{Size}}\\
    \cline{3-4}  \cline{5-6} 
    & & \textbf{mAP}& \textbf{P@100}&\textbf{mAP} &\textbf{P@100} \\
    \hline
              &Siamese CNN \cite{qi2016sketch} &0.183 &0.143 &0.153& 0.122& 64 \\
              &SaN \cite{yu2017sketch} &0.129 &0.104& 0.112& 0.096& 512 \\
\textbf{SBIR} &3D Shape \cite{wang2015sketch} &0.070 &0.062 &0.063& 0.057 & 64\\
              &DSH (Binary) $\dagger$ \cite{liu2017deep} &0.171& 0.231 & 0.129& 0.189& 64 $\dagger$ \\ 
              &GN Triplet \cite{sangkloy2016sketchy} &0.204& 0.296  & 0.175& 0.253& 1024 \\ \hline
               & SSE  \cite{zhang2015zero} &0.154& 0.108 &0.133& 0.096 & 100\\
              &JLSE \cite{zhang2016zero} &0.131& 0.185 & 0.109& 0.155& 220 \\ 
     \textbf{ZSL}    &ZSH \cite{yang2016revisiting}&0.159 &0.214 & 0.141& 0.177& 64 \\
              &SAE \cite{kodirov2017semantic}&0.216& 0.293 & 0.167& 0.221& 300 \\ \hline
              &ZS-SBIR \cite{kiran2018zero} &0.196& 0.284 & 0.005& 0.001& 1024 \\ 

              &EMS \cite{lu2018learning}   & -& - & 0.259 & 0.369 & 512 \\
    \textbf{ZSL:SBIR}   
               & CVAE \cite{kiran2018zero}   & 0.225 & 0.333  & -& -& 4096  \\
              &SEM-PCYC \cite{dutta2019semantically} &0.349& 0.463 & 0.297& 0.426& 64 \\ 
               &SAKE \cite{sake2019} \footnote{SAKE uses the ImageNet data as auxiliary information during the training process which greatly helps in boosting up the performance of the network. Hence, direct comparison with this framework is not fair. Further details are given in the discussion section.} & 0.547& 0.692 & 0.475& {0.599}& 64 \\ %
               &\textbf{CrossATNet} & {0.413}&{0.487}  & {0.327}& {0.427} & {64} \\ \hline
               &ZSIH $\dagger$  \cite{shen2018zero} & 0.258 & 0.342  & 0.223& 0.294& 64$\dagger$ \\
 \textbf{Hashed ZSL:SBIR} & EMS $\dagger$ \cite{lu2018learning}  & -& -&  0.165 & 0.252 & 64$\dagger$ \\
                &SEM-PCYC$\dagger$ \cite{dutta2019semantically}&0.344& 0.399 & 0.293& {0.392} & 64$\dagger$ \\
                &SAKE \cite{sake2019}$\dagger$ \footnotemark[\value{footnote}]& 0.364& 0.487 & 0.359& {0.481}& 512 $\dagger$ \\ 
               &\textbf{CrossATNet} $\dagger$ & {0.365} & {0.411}  &{0.316} & 0.404 & {64}$\dagger$ \\
               \hline
\end{tabular}
\end{center}
\end{minipage}
}
\end{table*} 

Similarly, the TU-Berlin dataset~\cite{liu2017deep} also has multiple unpaired sketch and image instances. There are a total of 250 classes in this dataset. Eight hundred images in each class and a variable number of sketches per class. The total number of sketch instances over 250 classes is 204,489. For our ZSL framework, we randomly select 30 classes as the unseen classes for the inference phase and use the remaining 220 for the training purpose as the training dataset.  

\noindent \textbf{Model Architecture:} To train the stage I of our network, we use the standard VGG-16, ResNet-50, and ResNet-101 pre-trained model and transfer the knowledge to our dataset. The model is then fine-tuned by minimizing a cross-entropy loss function to boost the descriptiveness of the features. These weights are then used for the initialization of stage II. While we test the performance of our network on three different pre-trained models, we report the final performance with the VGG-16 model to maintain fairness in the comparative study with the literature. The pre-trained network yields a 2048-d vector. We train the system using a momentum optimizer as it helps to accelerate SGD in the relevant direction and dampens oscillations. We choose a learning rate of 0.001 heuristically.

To train the Stage II of our network, we train the semantic encoder first. For the text-based model, we extract the 300-d word2vec features of the label names of the datasets. For the graph-based model, we created a $|\mathcal{Y}^s|$ dimensional distance matrix. The distance matrix was constructed by finding the Euclidean distances between the Word2Vec embedding of each class. A hierarchy-graph was constructed by taking the minimum spanning tree (MST) of this edge matrix. After obtaining the graph structure, we performed one layer of graph convolution. An auto-encoder was then used to combine the graph-based and the text-based models and construct a powerful reduct, capable enough of having a discriminative knowledge power (i.e., $f_w()$). The auto-encoder is designed to bring down the dimension of the feature vector to $64$. 

To realize the visual encoder part of our network, we use a series of convolution neural network layers, followed by a fully-connected layer for both $f_x()$ and $f_y()$. For the cross-attention module, we branch out a network from the sketch encoder and subject it to a global average pooling layer. This is then followed by two consecutive fully-connected layers and a sigmoid function to mask out the attention part. We then take the scalar product of this output and the visual encoding from the image branch $f_y(\theta_y)$ and use that as the new $f_y(\theta_y)$. 

For the decoder part, we use a single fully-connected layer placed after the last layer of the visual encoders. We put a batch normalization layer after the CNN layers and a drop out layer after the fully connected layers. We have also induced non-linearity into the network by placing \textit{leaky_ReLU()} layers after the drop-out functions. For the second stage, we use an Adam optimizer to minimize the loss function using a  stochastic gradient descent procedure. Again, the learning rate is heuristically chosen as 0.001. The model was trained by setting the $\alpha$ value as 1 (equation~\ref{eq:3a}) and a batch size of 256 for 200 epochs.

\noindent \textbf{Training and Evaluation Protocol:} 
We train the network by selecting random images and sketches and forming their corresponding triads. We train the network by approximately using 2,00,000 sketch-anchored and image-anchored triads. To avoid any training bias due to imbalance number of samples in the two modalities, we feed an equal number of triads of both types in each batch. To evaluate the performance of our model, we use the standard mean Average Precision (mAP) value and the P@100 (precision for top-100) scores and follow the training and evaluation protocol of~\cite{dutta2019semantically}. For the ZSL:SBIR part, we use just the sketch-anchored triplets to have a fair comparison with the state-of-the-art methodologies. Such training boosts the SBIR part of the performance of our network. For the cross-modal retrieval part, we train the model with both the sketch-anchored and image-anchored cross triplets.

\section{Discussions}
The train and test classes were chosen randomly for the experiments to avoid any bias induced while training. Table~\ref{tab:sketchyberlin} shows the comparison of the performance of various models present in the literature on the TU-Berlin and the Sketchy dataset. The current state-of-the-artwork in literature closest to our work is given in~\cite{dutta2019semantically,sake2019}. We also show a few more comparison with~\cite{qi2016sketch,yu2017sketch,wang2015sketch,liu2017deep,zhang2018generative,sangkloy2016sketchy,zhang2015zero,zhang2016zero,yang2016revisiting,kodirov2017semantic,kiran2018zero,lu2018learning,shen2018zero}. Here, $\dagger$ represents the algorithms that have a hashed vector-representation.  We report the performance of our model for both cross-modal and uni-modal retrieval in table~\ref{tab:unicross}. 

\begin{table}\caption{{Performance of the proposed CrossATNet framework for cross-modal retrieval on the Sketchy and TU-Berlin datasets in terms of mAP and precision at top-100 (P@100) values. Here, $\dagger$ represents the algorithms that have a hashed vector-representation. }}\label{tab:unicross}
\begin{center}
\scalebox{0.95}{
\begin{tabular}{@{\extracolsep{2pt}} l llll@{}}
    \hline
    \multirow{1}{*}{\textbf{Task}} &
    \multicolumn{2}{c}{\textbf{Sketchy}} &
    \multicolumn{2}{c}{\textbf{TU-Berlin}}\\
    \cline{2-3}  \cline{4-5} 
    & \textbf{mAP}& \textbf{P@100} & \textbf{mAP} &\textbf{P@100} \\
    \hline
               Sketch$\rightarrow$Image & {0.372} & {0.430}  & {0.302}& {0.322} \\
               Sketch$\rightarrow$Image  $\dagger$  &0.347& 0.410 & 0.312 & 0.376 \\ \cline{1-1}
               Image$\rightarrow$Sketch & 0.359 & 0.312 & 0.289 & 0.298 \\ \cline{1-1}
               Sketch$\rightarrow$Sketch & 0.318 & 0.343 & 0.238 & 0.257 \\  \cline{1-1}
               Image$\rightarrow$Image & 0.561 & 0.594 & 0.465 & 0.466 \\ \hline
\end{tabular}}
\end{center}
\end{table}

For the non-hashing case, SAKE exceeds the performance of CrossAtNet for the majority of the cases, while for hashing case, the results are almost comparable. However, our framework can exceed the performance of all the generative ZSL:SBIR and the majority of discriminative ZS-SBIR approaches. It is visible that while for the non-hashing case, SAKE gets the better of CrossAtNet for the majority of the cases and for the hashing case, we outperform SAKE on the Sketchy dataset. The significant distinctions between SAKE and other zero-shot SBIR frameworks that are responsible for the success of SAKE are (i) SAKE uses the entire ImageNet data as auxiliary information during training to combat the catastrophic forgetting, which greatly helps in generalization to unseen-class samples in ZS-SBIR. Precisely, SAKE uses the rich semantic space of Imagenet and deploys a domain adaptation paradigm to ensure that the sketch-image data in ZS-SBIR comply with the semantic topology of Imagenet. However, this additional information is seldom used in the zero-shot learning literature~\cite{dutta2019semantically}, and we also have not used the same in CrossAtNet. (ii) SAKE uses a conditional auto-encoder based shared feature extractor coupled with ResNet-50 (CSE-ResNet50) for the sketches and photos. This definitely contributes to boosting the performance of SAKE.  In spite of this, we see that standard VGG-16 based CrossAtNet achieves superior results to the literature and even outperforms SAKE in a few cases.

\begin{figure} 
   \centering 
   \includegraphics[scale=0.13]{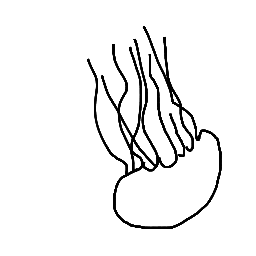}
   \fcolorbox{green}{yellow}{\includegraphics[scale=0.13]{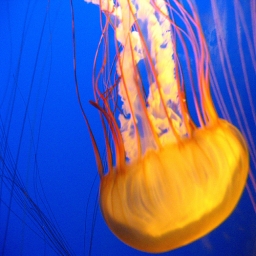}}
   \fcolorbox{red}{yellow}{\includegraphics[scale=0.13]{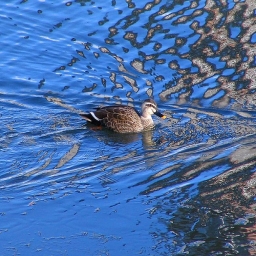}}
   \fcolorbox{green}{yellow}{\includegraphics[scale=0.13]{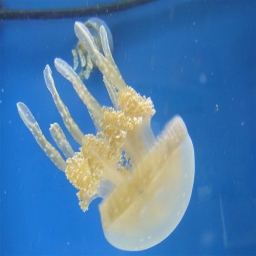}}
   \fcolorbox{green}{yellow}{\includegraphics[scale=0.13]{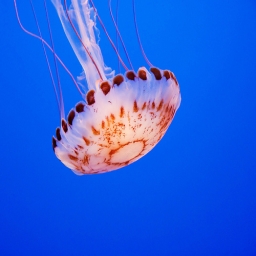}}
   \fcolorbox{green}{yellow}{\includegraphics[scale=0.13]{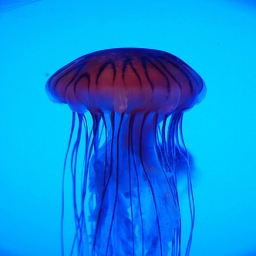}}\\
   \includegraphics[scale=0.13]{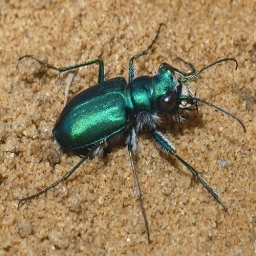}
   \fcolorbox{green}{yellow}{\includegraphics[scale=0.13]{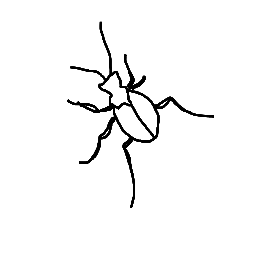}}
   \fcolorbox{green}{yellow}{\includegraphics[scale=0.13]{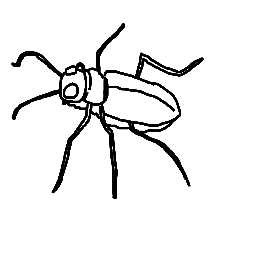}}
   \fcolorbox{red}{yellow}{\includegraphics[scale=0.13]{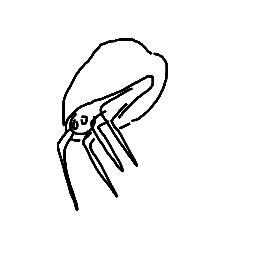}}
   \fcolorbox{green}{yellow}{\includegraphics[scale=0.13]{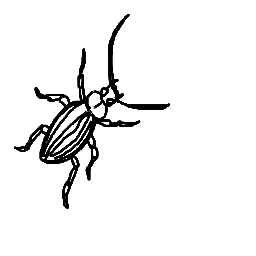}}
   \fcolorbox{green}{yellow}{\includegraphics[scale=0.13]{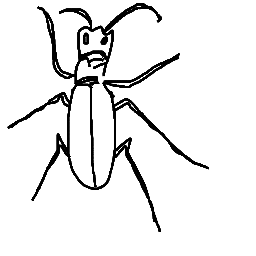}}\\
   \includegraphics[scale=0.13]{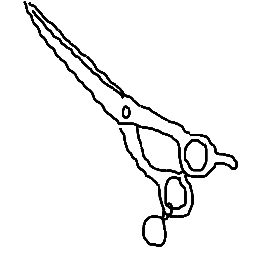}
   \fcolorbox{green}{yellow}{\includegraphics[scale=0.13]{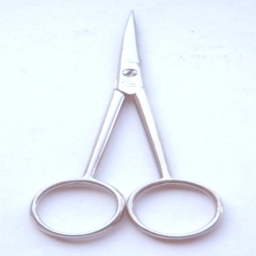}}
   \fcolorbox{green}{yellow}{\includegraphics[scale=0.13]{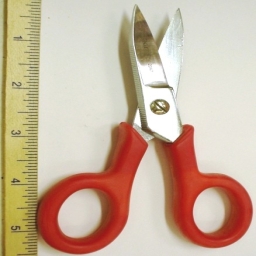}}
   \fcolorbox{green}{yellow}{\includegraphics[scale=0.13]{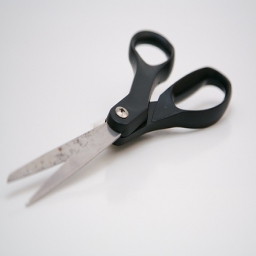}}
   \fcolorbox{green}{yellow}{\includegraphics[scale=0.13]{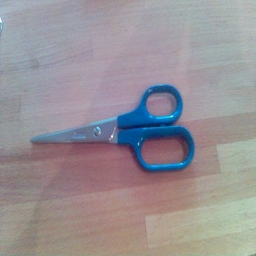}}
   \fcolorbox{green}{yellow}{\includegraphics[scale=0.13]{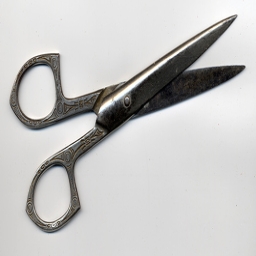}}\\
   \includegraphics[scale=0.13]{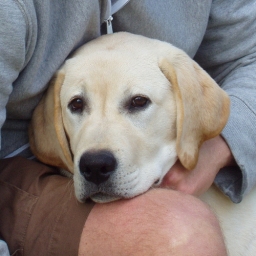}
    \fcolorbox{green}{yellow}{\includegraphics[scale=0.13]{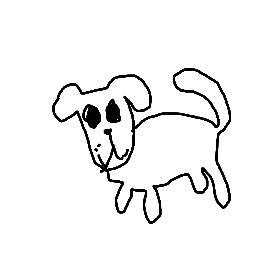}}
   \fcolorbox{red}{yellow}{\includegraphics[scale=0.13]{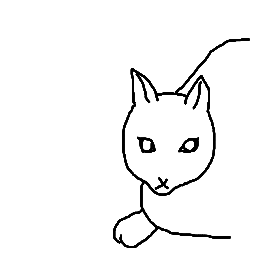}}
   \fcolorbox{green}{yellow}{\includegraphics[scale=0.13]{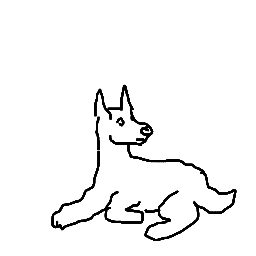}}
   \fcolorbox{red}{yellow}{\includegraphics[scale=0.13]{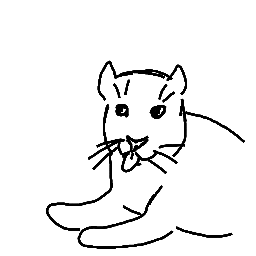}}
   \fcolorbox{green}{yellow}{\includegraphics[scale=0.13]{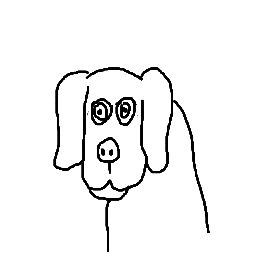}}\\

    \caption{Top retrieved results of zero-shot cross-modal retrieval on the unseen classes. Alternate rows represent Sketch$\rightarrow$Image and Image$\rightarrow$Sketch retrievals.}
    \label{fig:retrieved} \vspace{-2mm}
\end{figure}

\begin{figure}
    \centering
\subfloat[Images]{  \includegraphics[width=0.45\linewidth]{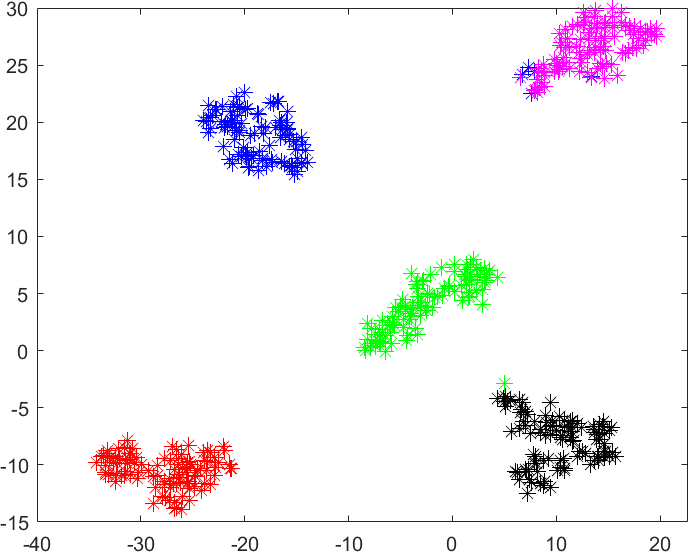}}
    \subfloat[Sketches]{\includegraphics[width=0.45\linewidth]{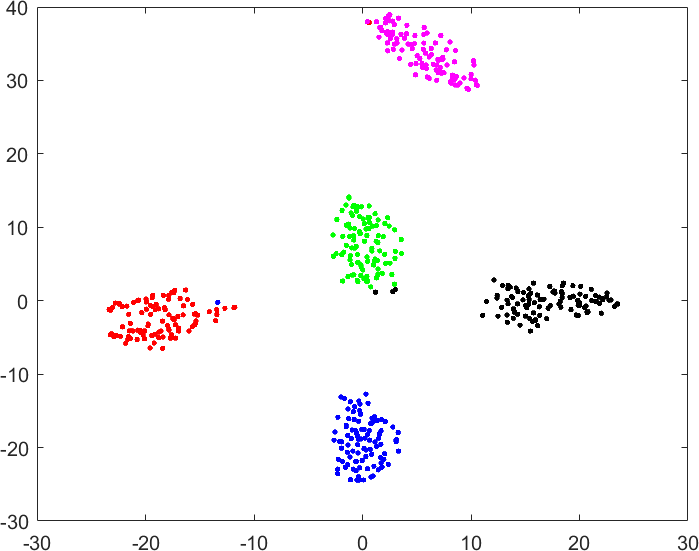}}\\
    \subfloat[Images and Sketches]{\includegraphics[width=0.45\linewidth]{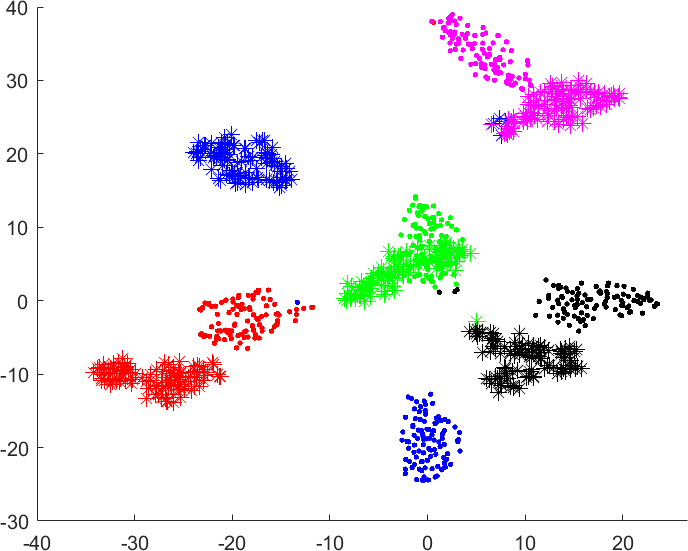}}
    \caption{Two-dimensional $t$-SNE scatter plots of the shared feature space of images, sketches, and both of them together for five random unseen classes. The different colors represent the embeddings of different classes. The image embeddings are plotted by hashes, while the sketch embeddings are plotted by dots.}
    \label{fig:tsne}\vspace{-3mm}
\end{figure}

The framework not only encodes the SBIR mapping, but it also encodes the inverse IBSR mappings. Table~\ref{tab:unicross} how well the proposed framework can retrieve images from sketches, sketches from images, images from images, and sketches from sketches.  Fig.~\ref{fig:retrieved} shows a few examples of the top-5 retrieved images, given a query image (first column). The two primary purposes of our framework were to make the shared feature space discriminative and to reduce the domain-gap between the two modalities. Fig.~\ref{fig:tsne} (a) and (b) shows the $t$-SNE plots of five random classes from the shared features of the images and features, respectively. It can be seen that the designed features space is discriminative. Fig.~\ref{fig:tsne} (c) shows the overlapped scatter plots of the two modalities to highlight how overlapping the feature space is. The image features are presented by crosses, while the sketch features are shown by dots.

\begin{figure}\center
  \begin{tabular}{ c c c c c }
   \includegraphics[scale=0.145]{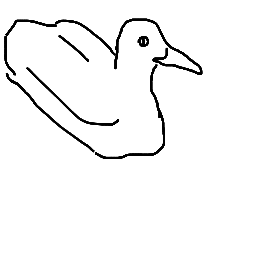}&
   \includegraphics[scale=0.145]{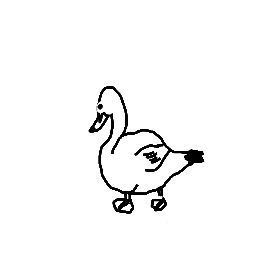}&
   \includegraphics[scale=0.145]{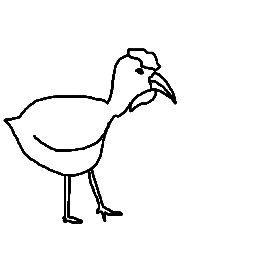}&
   \includegraphics[scale=0.145]{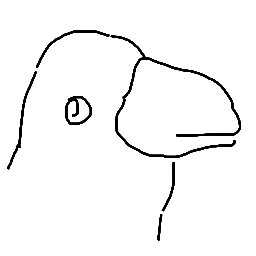}& 
   \includegraphics[scale=0.145]{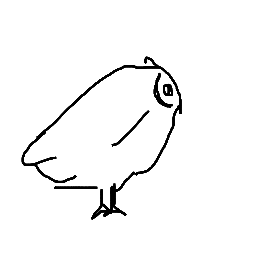}\\ \hline
   Duck & Swan& Chicken & Penguin& Owl\\ \hline
   \includegraphics[scale=0.145]{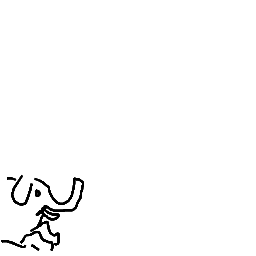}&
   \includegraphics[scale=0.145]{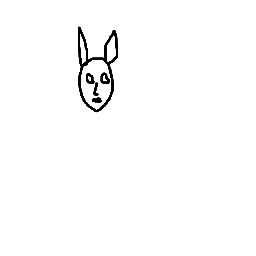}&
   \includegraphics[scale=0.145]{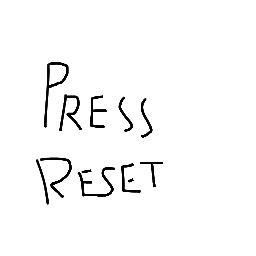}&
   \includegraphics[scale=0.145]{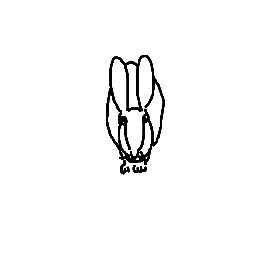}&
   \includegraphics[scale=0.145]{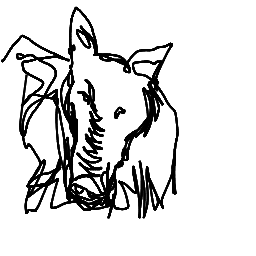}\\ \hline
    Elephant & Squirrel & Airplane & Rabbit& Pig\\ \hline
    \end{tabular}
  \caption{Few examples from confusion classes are represented in the first row. The second row shows some idiosyncratic sketches.}
  \label{fig:confusion}
\end{figure}

While performing the experiments, we observed that there is much confusion between very closely related classes, like {\tt Swan}, {\tt Duck}, {\tt Owl}, and {\tt Chicken}. The first row of Fig.~\ref{fig:confusion} shows a few instances from these classes. Upon close examination, we could observe that while the photos of these classes were well distinguishable, the sketches were tough to be recognized even with human evaluation. The main feature distinguishing them is the texture and colour properties, which obviously cannot be represented in sketches.  Also, we noticed that there were several idiosyncratically drawn sketches like shown in the second row of Fig.~\ref{fig:confusion}. There were multiple instances in each class, which just had ``press reset" written in it. Another interesting observation that we made during the critical analysis of the results was retrieving uni-modal photos was achieved with much higher precision than retrieving the uni-modal sketches. 
We presume that this is possibly due to the lack of highly informative texture data field.

\noindent \textbf{Hyper-parameter tuning of equation~\ref{eq:total}} ($\lambda_1, \lambda_2, \lambda_3$, and $\lambda_4$): We followed the standard procedure of the ZSL community by splitting the training data into (pseudo-seen and pseudo-unseen) for the purpose of cross-validating the parameters in equation~\ref{eq:total}. The network performance under these hyper-parameter settings in terms of mAP values on the Sketchy dataset is shown in table~\ref{tab:hyper}. We experiment the performance of the network by choosing different values of $\lambda$ ranging from 0.01 to 1. We choose the parameters $\lambda_1$ to $\lambda_4$ after cross-validation.  A sensitivity analysis on the parameters is shown in Table~\ref{tab:hyper}. We choose the combination of hyper-parameters that yield the best performance in the final model. 
\begin{table}
    \caption{Sensitivity analysis of the hyper-parameters for the final objective function by varying the values of the $\lambda_{i}$, ranging from 0.01 to 1. The results are reported in terms of the mAP values for the Sketchy dataset.}\label{tab:hyper}
    \begin{center}
        \begin{tabular}{c|cccc}\hline
        \textbf{Values}& \textbf{$\lambda_1$} & \textbf{$\lambda_2$} & \textbf{$\lambda_3$} & \textbf{$ \lambda_4$} \\ \hline
        0.01    &   0.36  &   0.37   &    0.32   &   0.41 \\
        0.1     &   0.41 &    0.41   &    0.29   &   0.39  \\
        1    & 0.11   &  0.19   &  0.41  &   0.38\\ \hline
    \end{tabular}
    \end{center}
\end{table}

\subsection{Ablation Studies}
\noindent \textbf{Effect of Graph-based model:} To investigate the contribution of graph structure in preserving the hierarchy of the labels in the semantic space, we study the performance of the proposed model with and without using the graph structure. We get the projection layer values just from the Word2Vec embeddings, followed by a set of auto-encoders to bring down the dimension of the projection vector. The leverage of preserving the hierarchy information of the labels can be seen from Table~\ref{tab:ablation}. 

\noindent \textbf{Effect of decoder loss ($\mathcal{L}_{rcs}$):} The decoder loss was introduced in the model to  achieve domain-independence between $\mathcal{X}$ and $\mathcal{Y}$. This loss reconstructs the feature embedding of the alternate modality from the shared-embedding of a given modality of data. Minimizing the decoder loss in addition to the overall loss function boosts the performance of the framework significantly. A definite fall in the performance of the network can be seen in table~\ref{tab:ablation}, without the decoder loss.

 {\noindent \textbf{Effect of cross-triplet loss ($\mathcal{L}_{3lt}$):} The cross-triplet was added to the network to bring the same class samples of different modalities nearby, while pushing the different classes of different modalities far apart in the embedding space. While performing a gradient descent on this loss function we can see a clear boost in the performance of this model from table~\ref{tab:ablation}. }
 
\noindent \textbf{Effect of fixed/latent semantic space:} To study the adequacy of the semantic space, we conduct two set of experiments. Firstly, we train the model with the Word2Vec features as the fixed semantic vector. In the second set of experiments, we keep this vector learnable and learn the optimum 300-d semantic vector using the Word2Vec and a graph hierarchy during the training process. It can be seen from table~\ref{tab:ablation} how the latent semantic vector outperforms the fixed projection vector.

\noindent \textbf{Effect of Pre-training:} To investigate the effect of bias introduced in the model due to the pre-training network, we perform our experiments using three different pre-trained models VGG-16, ResNet-50, and ResNet-101. It can be seen from table~\ref{tab:ablation} that a ResNet-50 trained model outperforms the VGG-16 trained model, while ResNet-101 on the other hand  performs a little inferior to ResNet-50 in the Sketchy dataset. For the TU-Berlin data, VGG16 yields better results than the other two. However, to keep the training protocol same for the comparative study, we report the VGG16 results in the table~\ref{tab:sketchyberlin}.

\begin{table}\caption{{Ablation study (with mAP values) with different experimental setups to analyze the effect of each of these dependencies on the proposed model. }}\label{tab:ablation}
\begin{center}
 \scalebox{0.9}{
\begin{tabular}{@{\extracolsep{2pt}}lccc@{}}
    \hline
    \multirow{1}{*}{\textbf{Task}} &
    \multicolumn{1}{c}{\textbf{Sketchy}}&
    \multicolumn{1}{c}{\textbf{TU-Berlin}} &
    \multicolumn{1}{c}{\textbf{dim}}\\
    \hline
     Total model & 0.413 & 0.327 & 64 \\ \cline{1-1}
    Without Graph  & 0.386 & 0.311 & 64 \\ \cline{1-1}
      {Without decoder loss $\mathcal{L}_{rcs}$}   & 0.215& 0.178 & 64\\  \cline{1-1}
      {Without triplet loss $\mathcal{L}_{3lt}$}   & 0.198& 0.217 & 64\\  \cline{1-1}
     Fixed semantic space  &0.243 &0.1987 & 300\\  \cline{1-1}
     Pretraining with VGG-16 & 0.413 & 0.327 & 64 \\  
     Pretraining with ResNet-50 & 0.421&0.325 & 64 \\
      {Pretraining with ResNet-101} & 0.398 & 0.309 &64 \\ \cline{1-1}
       {Seen+Unseen class graph} & {0.422} & {0.331}& 64 \\ 
      \hline
\end{tabular}
 }
\end{center}
\end{table}

{\noindent \textbf{Effect of graphs from seen and unseen classes:} In this experimental studies, we used two types of graphs while training. In first case we used the complete set of all classes to construct an initial minimum spanning tree from all the labels and learn the attribute space from it. In the second case, we just used the seen class labels to construct the graph and for the testing stage, we used this learning to get the attribute vectors. It is noted that using the complete graph yields better retrieval results than using just the seen class label graphs. However, in order to be fair and consistent with the other comparative algorithms used, we have shown our results with the one with just the seen class semantic space training.}

\section{Conclusion}
In this study, we proposed a novel deep representation learning technique for zero-shot sketch retrieval. The proposed framework is also robust to cross-modal and uni-modal retrieval set-up, using which we can also get the image-based sketch retrieval for the unseen classes. The network mainly brings closer the sample embeddings of different modalities in the shared space with the help of cross-triplets and an cross-attention network. By utilizing the novel cross-attention module and a well preserved semantic topography, we were able to beat the state-of-the-art models which exploit the standard protocol of a ZSL:SBIR framework. We also outperform the SAKE framework in a few cases, wherein the authors make use of auxiliary datasets to boost up their network performances considerably. The foremost essence is to design the shared embedding space sufficiently class-wise discriminative enough and adequately domain-wise invariant. We can extend the framework to form a self-supervised architecture or explore the possibility of extending it to an incremental learning framework.

\small{
\bibliography{mybibfile}
}
\end{document}